\documentclass{article}
\usepackage[nonatbib, final]{Styles/neurips_2024}
\RequirePackage[numbers]{natbib}

\usepackage{microtype}
\usepackage{graphicx}
\usepackage{subfigure}
\usepackage{booktabs} 

\usepackage{tabularx}
\usepackage{multirow}
\usepackage{wrapfig}
\usepackage{colortbl}

\usepackage{hyperref}
\usepackage{url}

\usepackage{xspace} 

\usepackage{amsmath}
\usepackage{amssymb}
\usepackage{mathtools}
\usepackage{amsthm}

\usepackage[capitalize,noabbrev]{cleveref}

\theoremstyle{plain}

\theoremstyle{definition}

\theoremstyle{remark}

\usepackage[textsize=tiny]{todonotes}

\definecolor{Gray}{gray}{0.90}

\def\modelname{BLIP-3-Video\xspace}

\renewcommand{\cite}{\citep}

\title{xGen-MM-Vid (\modelname): You Only Need 32 Tokens to Represent a Video Even in VLMs}

\author{%
  \textbf{Michael S. Ryoo$^1$, Honglu Zhou$^1$, Shrikant Kendre$^1$, Can Qin$^1$, Le Xue$^1$, Manli Shu$^1$,}\\
  \textbf{Jongwoo Park$^2$, Kanchana Ranasinghe$^2$,}\\
  \textbf{Silvio Savarese$^1$, Ran Xu$^1$, Caiming Xiong$^1$, Juan Carlos Niebles$^1$}\\
  $^1$Salesforce AI Research ~~~~~~~~~~~~~~~ $^2$Stony Brook University\\
  \texttt{mryoo@salesforce.com} \\
}

\begin{document}

\maketitle

\begin{abstract}
We present xGen-MM-Vid (\modelname): a multimodal language model for videos, particularly designed to efficiently capture temporal information over multiple frames. \modelname takes advantage of the `temporal encoder' in addition to the conventional visual tokenizer, which maps a sequence of tokens over multiple frames into a compact set of visual tokens. This enables \modelname to use much fewer visual tokens than its competing models (e.g., 32 vs. 4608 tokens). We explore different types of temporal encoders, including learnable spatio-temporal pooling as well as sequential models like Token Turing Machines. We experimentally confirm that \modelname obtains video question-answering accuracies comparable to much larger state-of-the-art models (e.g., 34B), while being much smaller (i.e., 4B) and more efficient by using fewer visual tokens.
\end{abstract}

\section{Introduction}

Large Vision-Language Models (VLMs), benefiting from large-scale image-text training, have been dominating the field of computer vision.
Recently, open-source VLMs are obtaining strong results, despite having much smaller size than the commercial models (e.g., 4B vs. Trillions).

\begin{wrapfigure}{r}{0.44\textwidth}
\vspace{-12pt}
    \centering
    \includegraphics[width=0.92\linewidth]{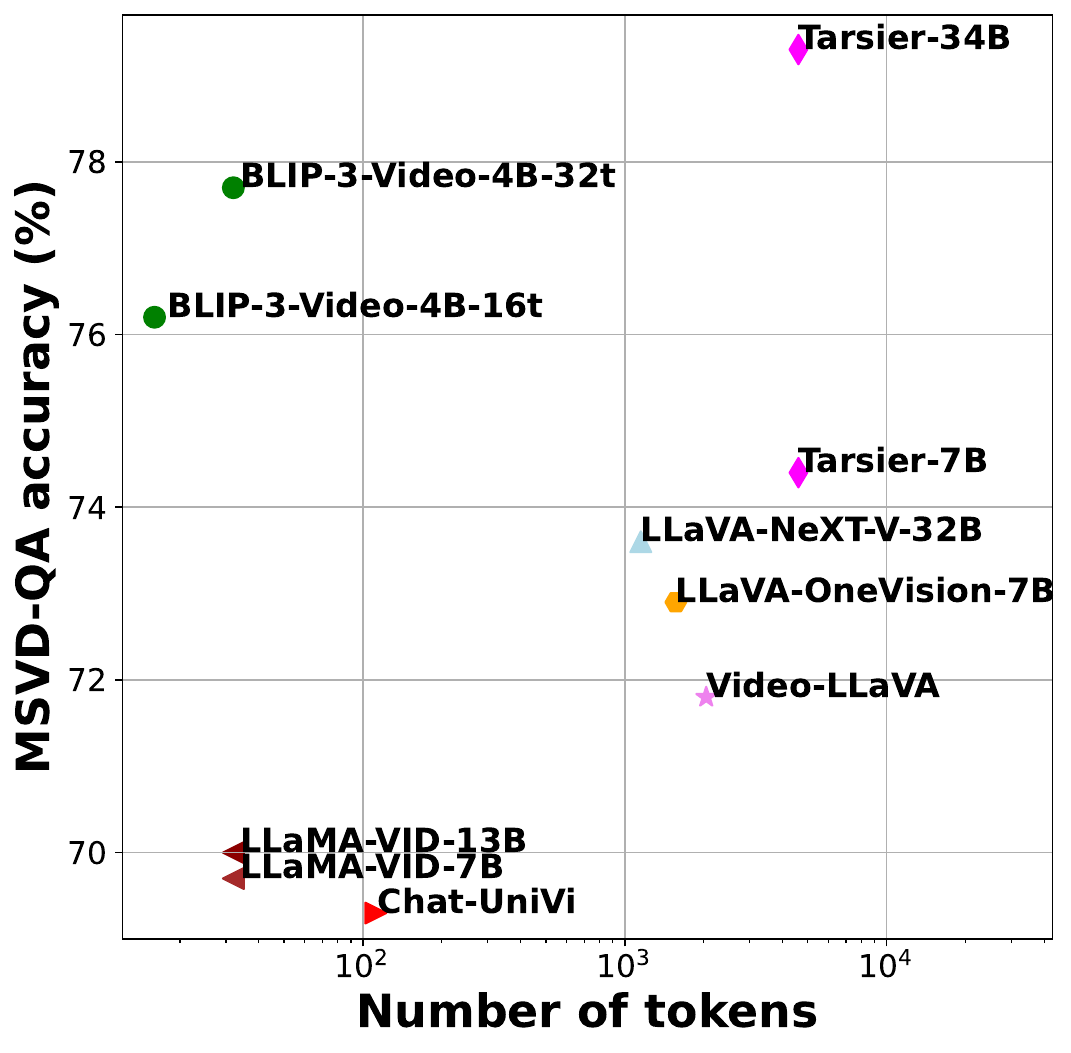}
    \caption{SOTA video VLM model comparison: Number of visual tokens vs. video-QA accuracy.}
    \vspace{-10pt}
    \label{fig:tradeoff}
\end{wrapfigure}


In addition to VLMs trained with images, VLMs for videos are becoming increasingly popular. The key component in a VLM for videos is the temporal abstraction of tokens over multiple frames. Models like Video-ChatGPT \cite{Maaz2023VideoChatGPT} and PLLaVA \cite{xu2024pllavaparameterfreellava} rely on a simple spatial/temporal pooling on top of image frame-level tokens to represent the entire video. Some models rely on a separate video encoder to capture temporal information in videos \cite{lin2023video}. Similarly, some models use additional convolutional layers (or Transformer layers) over frames to reduce their representation size (e.g., Video-LLaMA \cite{zhang2023video}, Kangaroo \cite{liu2024kangaroo}). Approaches that simply collect all the visual tokens from all the frames (e.g., MiniGPT4-video \cite{ataallah2024minigpt4}, LLaVA-NeXT \cite{li2024llavanext-interleave}, Tarsier \cite{wang2024tarsier} and LLaVA-OneVision \cite{wang2024tarsier}) also have been very popular recently, as they allow capturing all the details from the frame-level tokens. 

However, this often makes the number of tokens for video to be very huge (e.g., thousands even for 8 frames). Such large number of video tokens could be critical for longer videos as the LLM computation is quadratic to the number of total tokens.

\begin{figure*}[t]
    \centering
    \includegraphics[width=0.95\linewidth]{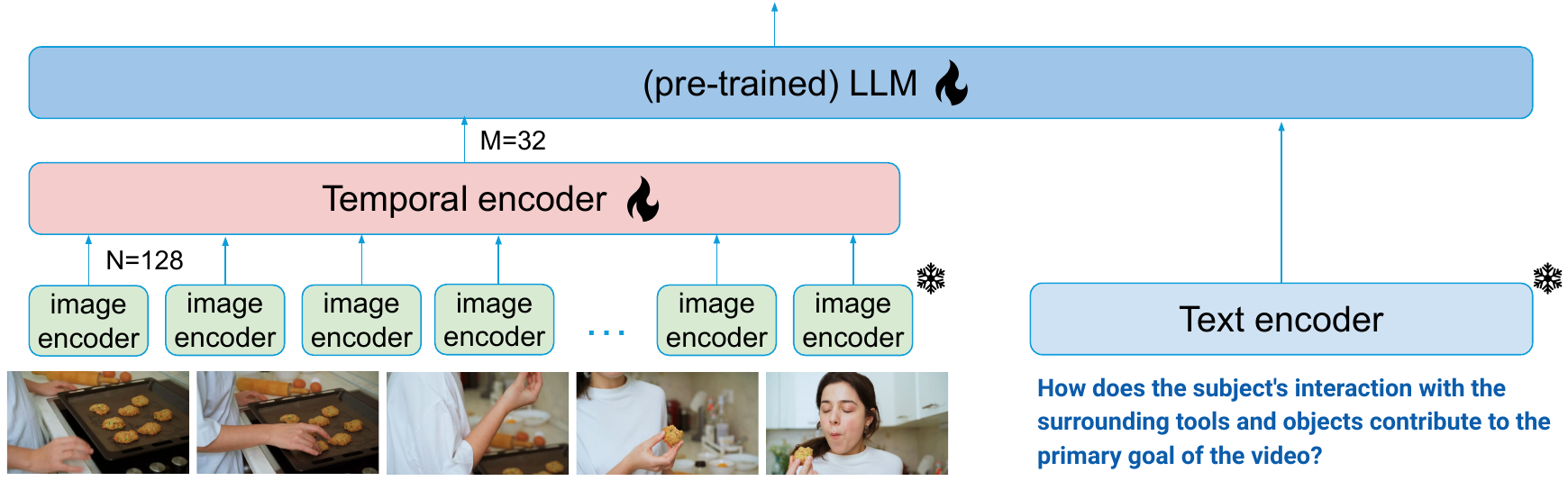}
    \vspace{-0.5em}
    \caption{\modelname model architecture. It has an explicit temporal encoder inserted to BLIP-3.}
    \label{fig:arch}
    \vspace{-0.5em}
\end{figure*}

In this paper, we introduce \modelname, which is an efficient compact vision-language model with an explicit \emph{temporal encoder}, designed particularly for videos. \modelname particularly focuses on incorporating a learnable `temporal encoder' within it. We explore different types of temporal encoder, and demonstrate that the model can abstract each video into much fewer visual tokens (e.g., 16) while being successful in open-ended question-answering and captioning tasks. We include a space-time attentional pooling as well as a sequential model as our temporal encoder, relying on token operations to iteratively abstract a series of frame-level tokens into a learnable memory. 


There has been prior work investigating the role of pooling \cite{jin2024chat}, convolutions, and cross attention layers \cite{zhang2023video,liu2024kangaroo,li2024llamavid}, but study on full space-time attentional pooling or sequential model to this extent has been limited in the past. Our objective in this paper is to provide a fundamental alternative to more brute-force way of collecting all the visual tokens which have been increasing popular recently. We experimentally confirm that $16\sim32$ video tokens abstracted by the temporal encoder is often sufficient to represent the entire video for question-answering (Figure \ref{fig:tradeoff}).


\section{\modelname}

\subsection{Model architecture}

We build \modelname based on the image-based vision-language model (VLM), BLIP-3 \cite{xue2024xgenmmblip3}.
The model architecture is composed of the following four components: (1) the vision encoder (ViT) taking each frame input, (2) the frame-level tokenizer to reduce the number of tokens, (3) the temporal encoder to build video-level token representations, and (4) the autoregressive LLM generating output text captions based on such video tokens and text prompt tokens. Figure \ref{fig:arch} shows an overview.

First, we apply a pretrained SigLIP as the vision encoder, designed to take one single image frame at a time. Perceiver-Resampler is then applied to map such visual tokens into $N=128$ visual tokens per frame, independently.
Once the model has such visual tokens over time (i.e., over multiple frames in the video), they are provided to an explicit `temporal encoder'. The role of the temporal encoder is to build a video-level token representation from such sequence of image-level tokens, serving as a mapping function between a set of $N \cdot T$ image tokens to $M$ video tokens where $T$ is the number of frames and $M$ is a constant number of tokens. We explore various forms of the temporal encoder, including temporal pooling as well as sequential models, which we discuss further in the following subsection. The resulting tokens are given to the LLM together with the encoded text tokens in a prefix manner, as in many standard VLMs.

For computational efficiency, the model takes uniformly sampled 8 frames per video. As a result, in our model, ViT first maps a video into $8 * 729$ visual tokens, which is then mapped to $8 * 128$ visual tokens using Perceiver-Resampler, and then to $16\sim128$ video tokens using the temporal encoder.

We use Phi-3 \cite{abdin2024phi3} as our LLM backbone taking such video tokens in addition to the text prompt tokens. This enables the model to take text+video as an input and generate text sentences as an output.

\begin{figure*}
    \centering
    \includegraphics[width=0.85\linewidth]{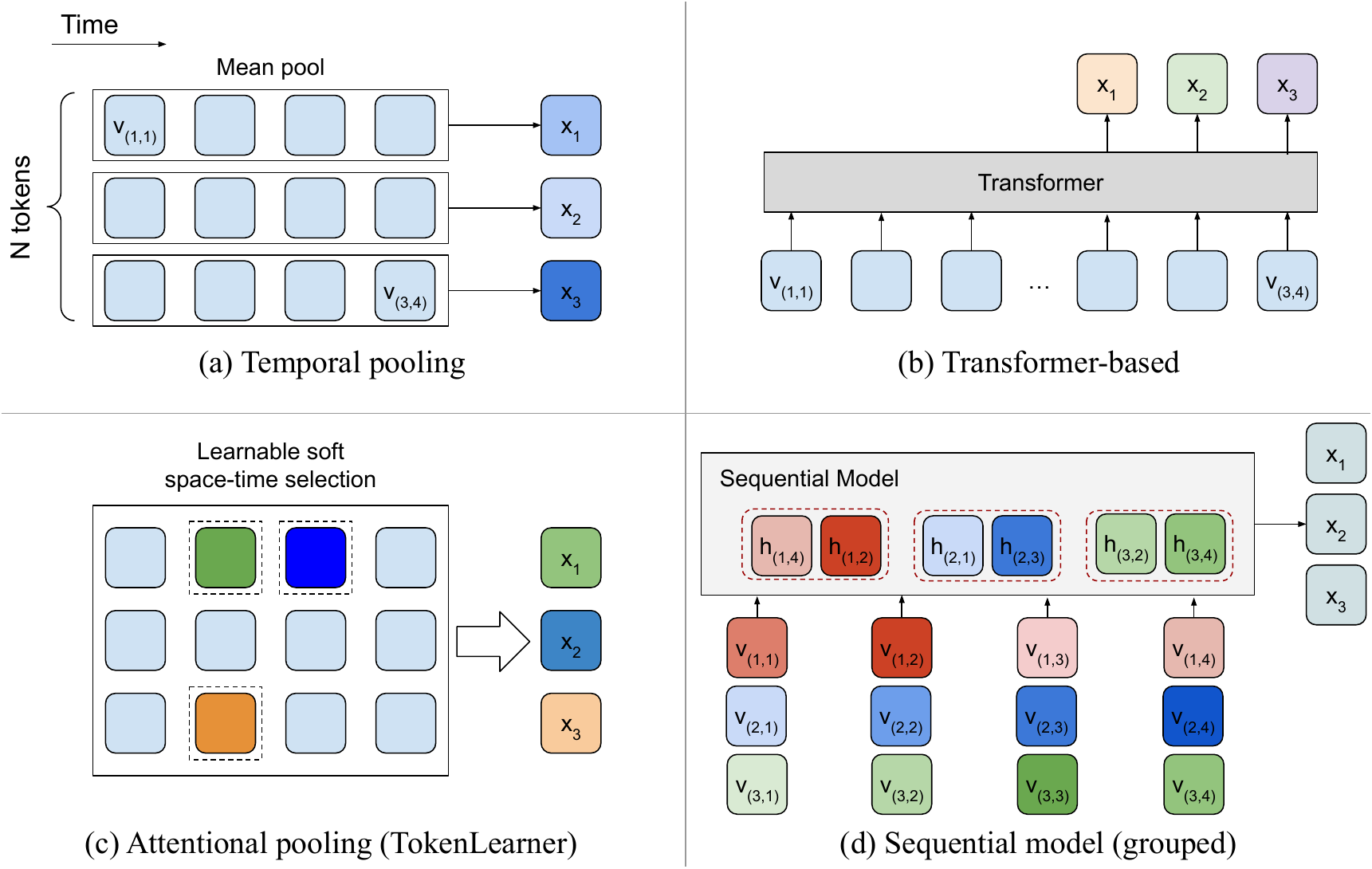}
    \caption{Visually comparing different types of temporal encoders we explored in our model architecture. (c) and (d) are particularly effective, as we discuss further in the experiments.}
    \label{fig:temporal}
\end{figure*}

\subsection{Temporal encoders}

A temporal encoder is a function of tokens, taking $N \cdot T$ tokens as an input and returning $M$ tokens as an output: $x_{1,\ldots, M} = f(v_{(1, 1), \ldots, (N, T)})$.
We explore different types of encoders as part of our model. The simplest form of the temporal encoder will be temporal pooling, e.g., summating per-frame tokens over time: $x_{1,\ldots, M} = \left\{\sum_t (v_{(j, t)}) \right\}_{j=1}^M$ where $M$ is  restricted to be identical to $N$, which was also used in \cite{Maaz2023VideoChatGPT}. Another possible implementation would be the use of a temporal Transformer, modeling the entire token sequence and selecting the last $m$ tokens as in Mirasol3B \cite{piergiovanni2024mirasol3b}:
\begin{align}
    x_{1,\ldots, M} = \left\{\textrm{Transformer}(v)\right\}_{N \cdot T-M+1}^{N \cdot T}
\end{align}

In addition to these temporal encoders, we explore two important temporal encoders considering space-time nature of tokens: spatio-temporal attentional pooling and sequential models (Figure \ref{fig:temporal}).

\noindent\textbf{Spatio-temporal attentional pooling:}
Attentional pooling allows learnable `soft selection' of multiple tokens given a larger set of tokens. It was developed for Transformers (e.g., Perceiver \cite{jaegle2022perceiver} and TokenLearner \cite{ryoo2021tokenlearner}), and also used in earlier foundation models (e.g., CoCa \cite{yu2022coca}) for images.
In our model, we use TokenLearner~\cite{ryoo2021tokenlearner}, which explicitly serves as our space-time aware temporal encoder. Unlike previous per-frame use of pooling where spatial pooling and temporal pooling are applied separately \cite{Maaz2023VideoChatGPT}, our temporal encoder directly takes all $N \cdot T$ tokens and `learns' to soft-select $M$ informative tokens spatio-temporally. Here, $N$ tokens could be viewed as spatial representations of a frame and we have $T$ of them, suggesting it is a spatio-temporal representation selection.

Our attentional pooling in its simplest form is expressed as:
\begin{align}
    x_i = A(V) \cdot V = \textrm{softmax}\left(\alpha(V^T)\right) \cdot V
\end{align}
where $V$ is a matrix formed by concatenating input tokens $v_{(1, 1), \ldots, (N, T)}$. The function $A(\cdot)$ computes the summation weights for $V$, performing soft selection of tokens. This is further decomposed to the softmax and the function $\alpha(\cdot)$. In Perceiver, a matrix multiplication with a latent query tokens (i.e., cross attention where $|Q| = m$) have been used to implement this: $\alpha(V) = Q \cdot V^T / c$. TokenLearner uses a convolution/MLP on top of $V$: $\alpha(V) = \textrm{MLP}_m(V^T)$, which we use in our model. This allows selecting a smaller number of tokens (e.g., $M=32$ tokens).

We experimentally confirm that such learnable spatio-temporal attentional pooling has advantages over the conventional approach of non-learnable spatial pooling and temporal pooling, in Section \ref{sec:ablations}.

\subsection{Grouped tokens sequential model}

Motivated by the success of sequential models in representing series of tokens such as Mamba \cite{gu2024mamba} and Token Turing Machines \cite{ryoo2023tokenturingmachines}, we design a new class of sequential models for videos and take advantage of it as our temporal encoder. The sequential models are capable of taking any number of inputs to generate a fixed number of output tokens, making it suitable to build a video-level token representation (e.g., $M=32$ regardless the number of frames).

The main distinction between our new sequential models and the conventional sequential models is that our sequential model maintains a grouped memory, separately processing and maintaining different visual features. It is in a way analogous to the operational difference in the standard convolution vs. the grouped convolution, but it is done not over channels but over tokens and they are processed with multiple time steps.

Let $F$ be the main model function in a standard sequential model: $h_i = F(h_{i-1}, v_i)$ where $h_i$ is the memory token and the index $i$ ranges from 1 to $N \cdot T$ (i.e., the total number of tokens) in the entire video. In the new grouped formulation, we instead maintain the set of tokens at time step $t$ as:
\begin{align}
H_t = \bigl\{F(h_{(j, t-1)}, v_{(j, t)}) \bigr\}_{j=1}^N
\label{eq:seq}
\end{align}
where $j$ is the visual token index within each frame. That is, we enforce the sequential model to focus on its group (specified with $j$) and maintain a `set' of memory tokens for every time step $t$: $H_t$.

We tried both the grouped version of Mamba and TTM in our implementation as temporal encoder architectures, finding that our grouped tokens sequential model based on TTM functions better. In its instantiation, we first extend TTM by adding time-stamped positional encodings to embed the frame index of each token in the latent space. This enables the tokens in the `memory' of TTM to preserve the temporal ordering information, which we found to be crucial when representing complicated or long video scenes.

Next, we implement Equation \ref{eq:seq}, while using the Read and Write operations of the original TTM. This follows our new grouped sequential model formulation, and the TTM now maintains a separate memory of size $G=4$ for each of $N=128$ tokens over time. That is, it maintains a `grouped' memory, which we found to better preserve scene details. The memory is maintained to have the size of $N \cdot G$, and the final output from the sequence model is attentionally pooled from the final memory to give $M$ tokens.



In our experiments, we confirm that the proposed class of sequential models perform significantly better than the conventional sequential models like TTMs.

\subsection{Training recipe}
\modelname follows a three-stage curriculum learning: (1) image caption pretraining, (2) video caption pretraining, and (3) video instruction tuning. In all its training we freeze the vision encoder, only training the model parameters after the vision encoder. First, we directly use the pretrained weights from BLIP-3 \cite{xue2024xgenmmblip3}. BLIP-3 is for images and it does not contain weights for the temporal encoder, so we randomly initialize those weights.

As its 2nd stage, the model is then trained on LLaVA-Hound-DPO's video caption data \cite{zhang2024direct}, featuring over 900k video captions. Instead of directly using the text captions provided in LLaVA-Hound-DPO, we used GPT-4 to rephrase such text captions so that they become more GPT-style captions. 


Finally, we tuned the model using a mix of video question-answering datasets, including VideoChatGPT's 99k-sample video instruction tuning data \cite{Maaz2023VideoChatGPT}, along with the training splits of the MSVD-QA~\cite{msvdqamsrvttqa}, MSRVTT-QA~\cite{msvdqamsrvttqa}, ActivityNet-QA~\cite{yu2019activitynet}, TGIF-QA~\cite{jang2017tgif}, and NExT-QA~\cite{xiao2021next} datasets, which contain 30k, 149k, 32k, 71k, and 34k samples, respectively. For TGIF-QA, we only used the training data associated with the Repeating Action and State Transition tasks. In our video instruction tuning recipe, we employ both open-ended and multiple-choice video QA formats for TGIF-QA and NExT-QA. For the open-ended video QA training data sourced from the MSVD-QA, MSRVTT-QA, TGIF-QA, and NExT-QA training sets, we used GPT-3.5 to rephrase the original single-word or single-phrase answer into a natural language sentence, providing the question in the LLM prompt context. For open-ended TGIF-QA and NExT-QA, we also double the sample size by using both the original short-phrase answers and the rephrased sentence-based answers.
In addition, we added a filtered version of the Mira caption dataset~\cite{ju2024miradatalargescalevideodataset} for our video instruction tuning. That is, we are using both video question-answering and video captioning for our final training. We excluded captions for Mira videos longer than one minute, totaling 935k video caption samples.

We trained our model with 8 $\times$ H100 GPUs. For the video caption pretraining, we use the batch size of 16 per GPU, 500 warmup steps, and the learning rate of 2e-5 with the cosine decay. We trained the model for 1 epoch. The video QA sft (i.e., instruction tuning) was done with the batch size of 4 per gpu, 500 warmup steps, and the learning rate of 1e-5  with the cosine decay. We trained the model for 1 epoch in this case as well. The entire training (combining both video pretraining and the sft) takes around 12 hours, confirming the efficiency of our model.



\section{Experiments and Results}

\noindent\textbf{Model implementation details.} We share the model details with BLIP-3 (4B), except that \modelname has the new temporal encoder component in its architecture. This model takes the video with the input resolution of 384$\times$384, using SigLIP encoder to map it to 729 tokens per frame with the channel size 1152. Perceiver-Resampler is implemented with multiple cross-attention layers with the same channel dim, which is then given to the temporal encoder.

TokenLearner serving as the spatio-temporal attentional pooling was implemented using a MLP as the attention function. The size of its inner dim was the number of target tokens * 2. The grouped sequential model temporal encoder was implemented using 4 Transformer layers (with the channel dim of 1152) as the processor module while using TokenLearners for read/write modules. Memory size was set to $N * 4 = 512$ tokens total.
The resulting $16\sim128$ tokens are mapped to the channel dim of 3072, before given to the LLM (Phi-3).

\begin{table*}[tb]
\caption{Comparison against reported numbers of other models on open-ended question answering evaluation. The number of visual tokens are also reported. The numbers after `/' are answer quality scores. $^*$ indicates our evaluation using the checkpoint and inference code provided by the author, with the identical videos used in our model (8 frames of 384$\times$384 resolution).}
\label{tab:comparison}
\centering
\small
\vspace{0.5em}
\def\arraystretch{1.1}  
\setlength\tabcolsep{0.6em}  
\scalebox{0.85}{
\begin{tabular}{l|r|r|r|r|r|r}
\hline
Method & Size & \#tokens & MSVD-QA & MSRVTT-QA & ActivityNet-QA & TGIF-QA\\
\hline
VideoChat~\cite{li2023videochat}  & 7B & \textbf{32} & 56.3 / 2.8  & 45.0 / 2.5 & - / 2.2 & 34.4 / 2.3 \\
Video-LLaMA~\cite{zhang2023video} & 7B & \textbf{32} & 51.6 / 2.5  & 29.6 / 1.8 & 12.4 / 1.1 & - / - \\
Video-ChatGPT \cite{Maaz2023VideoChatGPT} & 7B & 264+ & 64.9 / 3.3 &  49.3 / 2.8 & 34.2 / 2.8 & 51.4 / 3.0 \\
Chat-UniVi~\cite{jin2024chat} & 7B & 112 & 69.3 / 3.7 & 55.0 / 3.1 & 46.1 / 3.3 & 69.0 / 3.8\\
LLaMA-VID~\cite{li2024llamavid} & 7B & \textbf{32} & 69.7 / 3.7 & 57.7 / 3.2 & 47.4 / 3.3 & -\\
LLaMA-VID~\cite{li2024llamavid} & 13B & \textbf{32} & 70.0 / 3.7 &58.9 / 3.3 & 47.5 / 3.3 & -\\
Video-LLaVA \cite{lin2023video} & 7B & 2048 & 71.8 / 3.9 & 59.2 / 3.5 & 45.3 / 3.3 & 70.0 / 4.0\\
MiniGPT4-Video~\cite{ataallah2024minigpt4} & 7B & 2880+ & 73.9 / 4.1 & 59.7 / 3.3 & 46.3 / 3.4 & 72.2 / 4.1\\
PLLaVA \cite{xu2024pllavaparameterfreellava} & 7B & 576+ & 76.6 / 4.1 &  62.0 / 3.5 & 56.3 / 3.5 & 77.5 / 4.1\\
SlowFast-LLaVA~\cite{xu2024slowfast}  & 7B &  3680 & 79.1 / 4.1   &  65.8 / 3.6 &  56.3 / 3.4 &   78.7 / 4.2 \\  %
LLaVA-Hound-DPO~\cite{zhang2024direct}  & 7B &  2048 &  80.7 / 4.1  & 70.2 / 3.7 & - / -  &  61.4 / 3.5 \\
LLaVA-OneVision$^*$ \cite{wang2024tarsier} & 7B & 1568 &  72.9 / 3.9 &  57.8 / 3.4  &  55.3 / 3.6 & 41.1 / 3.1 \\
Tarsier \cite{wang2024tarsier} & 7B & 4608+ & 77.0 / 4.1 &  62.0 / 3.5 & 59.5 / 3.6 & 79.2 / 4.2\\
Tarsier $^*$ \cite{wang2024tarsier} & 7B & 4608 & 74.4 / 4.0  &  59.1 / 3.4  & 54.3 / 3.5  & - / - \\
\hline
PLLaVA \cite{xu2024pllavaparameterfreellava} & 34B & 576+ & 79.9 / 4.2 & 68.7 / 3.8 & 60.9 / 3.7 & 80.6 / 4.3\\
LLaVA-NeXT-Video$^*$ \cite{li2024llavanext-interleave} & 32B & 1152 & 73.6 / 4.0
  &  56.8 / 3.4  & 58.4 / 3.6
  &  73.5 / 4.1 \\
Tarsier \cite{wang2024tarsier}  & 34B & 4608+ & 80.3 / 4.2 & 66.4 / 3.7 & 61.6 / 3.7 & 82.5 / 4.4\\
Tarsier $^*$ \cite{wang2024tarsier} & 34B & 4608 & 79.3 / 4.1  &  62.2 / 3.5  & 61.5 / 3.7  &  - / -  \\
\hline
\modelname & \textbf{4B} & \textbf{32} & 77.7 / 4.2 & 60.0
/ 3.6 & 55.7 / 3.5 & 76.5 / 4.3 \\
\modelname & \textbf{4B} & 128 & 77.9 / 4.3 & 59.7 / 3.6 & 56.9 / 3.6 & 77.1 / 4.3\\
\hline
\end{tabular}
}
\vspace{-10pt}
\end{table*}

\subsection{Public benchmarks}

We conducted experiments measuring video question-answering accuracies on multiple public datasets. This includes open-ended answer generation tasks like MSVD-QA, as well as multiple choice questions like NExT-QA and MVBench \cite{mvbench2024}. We follow their standard settings in all cases.

Table \ref{tab:comparison} compare open-ended question answering accuracies of \modelname against reported numbers of other models. We use four commonly used public datasets, MSVD-QA, MSRVTT-QA, ActivityNet-QA, and TGIF-QA, following standard VideoLLM evaluation settings. Note that our MSVD-QA and MSRVTT-QA accuracy was measured by training our model with a subset (i.e., Video-ChatGPT dataset-only) of our training data, as this allows more direct comparison to some of the prior work and enables more stable results due to its data distribution. We are including the model size as well as the number of visual tokens in the table. We are able to observe that, despite its smaller size (i.e., 4B vs. 7B or 34B), our model is obtaining superior or comparable performance.

With the temporal encoder, \modelname was able to retain the performance with much fewer tokens, which we discuss more in the following subsection. Our results suggest that not too many visual tokens are really necessary to be successful on these open-ended question answering benchmarks, as long as we have an effective temporal encoder.

\begin{table}
\begin{minipage}{0.41\linewidth}
    \caption{Comparison against reported numbers of other models on NExT-QA.}
    \label{tab:nextqa}
    \centering
    \small
    \def\arraystretch{1.0}  
    \setlength\tabcolsep{0.5em}  
    \scalebox{0.95}{
    \begin{tabular}{l|r|r|c}
    \hline
    Method & Size & \#tokens & Acc. \\
    \hline
    LangRepo \cite{kahatapitiya2024langrepo} & 7B & 3136+ & 54.6\\
    LangRepo \cite{kahatapitiya2024langrepo} & 12B & 3136+ & 60.9\\
    Tarsier \cite{wang2024tarsier}  & 7B & 4608+ & 71.6\\
    \hline
    LLoVi \cite{zhang2024llovi} & 157B & 1000s & 67.7\\
    IG-VLM \cite{kim2024imagegridworthvideo} & 34B & 1536+ & 70.9\\
    VideoAgent \cite{wang2024videoagent} & GPT-4 & 2091+ & 71.3\\
    VideoTree \cite{wang2024videotree} & GPT-4 & 3978+ & 73.5\\
    Tarsier \cite{wang2024tarsier}  & 34B & 4608+ & 79.2\\
    \hline
    \modelname & \textbf{4B} & \textbf{32} & 76.4 \\
    \modelname & \textbf{4B} & 128 & 77.1 \\
    \hline
    \end{tabular}
    }
\end{minipage}
\hspace{0.02\linewidth}
\begin{minipage}{0.56\linewidth}
    \caption{Comparison on MVBench. `VC-IT' indicates whether the model was trained on the MVBench-provided training dataset. `$\sim$Y' means the model’s training recipe includes a major subset of VideoChat2-IT (e.g., CLEVRER, Kinetics-710, SthSthV2, WebVid, ...).}
    \label{tab:mvbench}
    \centering
    \small
    \def\arraystretch{1.0}  
    \setlength\tabcolsep{0.8em}  
    \scalebox{0.95}{
    \begin{tabular}{l|r|r|c}
    \hline
    Method (Size) & \#tokens & VC-IT & Acc. \\
    \hline
    PLLaVA (7B) &	576+ &	Y &	46.6\\
    VideoLLaMA2 (7B) &	1152 &	Y &	54.6\\
    ST-LLM (7B) &	256 &	$\sim$Y &	54.9\\
    PPLLaVA (7B) &	1024 &	$\sim$Y &	59.2\\
    VideoChat2-Mistral (7B) &	96 &	Y &	60.4\\
    Kangaroo (8B) &	$\sim$10000 &	Y &	61.1\\
    Tarsier (7B) &	4608+ &	$\sim$Y &	62.6\\
    \hline
    VideoChatGPT (7B) &	264+ &	N &	32.7\\
    VideoLLaMA (7B) &	\textbf{32} &	N &	34.1\\
    VideoChat (7B) &	\textbf{32} &	N &	35.5\\
    LLaMA-VID (7B) &	\textbf{32} &	N &	41.4\\
    Video-LLaVA (7B) &	2048 &	N &	43.5\\
    mPLUG-Owl3 (8B) &	n/a &	N &	54.5\\
    LLaVA-OneVision (7B) &	3136 &	N &	56.7\\
    \hline
    \modelname (\textbf{4B}) & \textbf{32} &	N &	54.9\\
    \hline
    \end{tabular}
    }
    \end{minipage}
\end{table}

In addition, we evaluated \modelname's ability to solve multiple choice questions (MCQ). Table \ref{tab:nextqa} shows the results on NExT-QA, and Table \ref{tab:mvbench} shows the results on MVBench. Due to the nature of its questions requiring understanding of multiple frames, many prior models use quite a bit of tokens. For instance, GPT-4 uses a minimum of 255 tokens per frame. It is interesting that \modelname achieves comparable accuracy while representing the entire video with only 32 (or 128) tokens.



\begin{table*}[tb]
\caption{Ablations comparing different temporal encoders: 128 tokens. $^*$A slightly different training recipe using a subset of the entire dataset (without Mira data) was used for the ablations.}
\label{tab:ablation128}
\vspace{0.5em}
\centering
\small
\def\arraystretch{1.0}  
\setlength\tabcolsep{0.9em}  
\scalebox{0.95}{
\begin{tabular}{c||c|c|c||c}
\hline
Encoder & MSVD-QA & TGIF-QA & ActivityNet-QA & NExT-QA\\
\hline
1 frame & 71.49 / 4.01 & 72.74 / 4.16 & 51.83 / 3.39 & 72.79 \\
Mean pooling & 76.75 / 4.17 & 77.01 /4.30 & 55.89 / 3.53 & 76.24 \\
Transformer & 76.24 / 4.15 & 76.33 / 4.28 & 55.59 / 3.50 & 76.34 \\
Perceiver-Resampler & 76.17 / 4.12 & 72.46 / 4.13 & 52.61 / 3.38 & 76.44 \\
Vanilla Token Turing Machine & 76.42 / 4.15 & 75.80 / 4.26 & 54.45 / 3.48 & 75.42 \\
\hline
Ours (Space-time) & 77.49 / 4.18 & 76.90 / 4.29 & 56.94 / 3.56 & 76.27 \\
Ours (Sequential) & 77.86 / 4.20 & 77.10 / 4.31 & 56.66 / 3.56 & 77.07 \\
\hline
\end{tabular}
}
\end{table*}

\subsection{Ablations}
\label{sec:ablations}

\noindent\textbf{Temporal encoder:}
We conducted an ablation comparing different temporal encoders within our model. These include: (1) the base single frame model (i.e., BLIP-3 trained with videos), (2) mean pooling similar to Video-ChatGPT, and (3) transformer temporal encoder similar to Mirasol3B. We also tried (4) the approach of directly extending the spatial Perceiver-Resampler to do spatio-temporal encoding, and (5) vanilla Token Turing Machines, which is not our grouped sequential model.

Table \ref{tab:ablation128} shows the result, comparing the question-answering accuracies of different types of temporal encoders when abstracting a video into 128 tokens. We are able to observe that our sequential model temporal encoder performs the best overall. In particular, directly extending Perceiver-Resampler performed poorly compared to the others.




\begin{table}[tb]
\parbox{.39\linewidth}{
\caption{Ablations comparing different pooling strategies (32).}
\label{tab:ablation32}
\centering
\small
\def\arraystretch{1.1}  
\setlength\tabcolsep{0.8em}  
\scalebox{0.95}{
\begin{tabular}{c|c}
\hline
Encoder & MSVD-QA \\
\hline
Space-time pooling (4*8) & 76.04 \\
Per-frame (4*8) & 76.78 \\
\hline
Ours (Space-time) & 77.07 \\
Ours (Sequential) & 77.11 \\
\hline
\end{tabular}
}
}
\hspace{0.02\linewidth}
\parbox{.58\linewidth}{
\caption{Ablations comparing different \# of tokens. Ours with sequential model as a temporal encoder was used.}
\label{tab:ablation-tokens}
\centering
\small
\def\arraystretch{1.0}  
\setlength\tabcolsep{0.7em}  
\scalebox{1.0}{
\begin{tabular}{c||c|c|c}
\hline
\# tokens & MSVD-QA & TGIF-QA & NExT-QA\\
\hline
16 tokens & 76.17 / 4.16 & 76.19 / 4.28 & 75.8\\
32 tokens & 77.11 / 4.17 & 77.07 / 4.30 & 76.4\\
128 tokens & 77.86 / 4.20 & 77.10 / 4.31 & 77.07\\
256 tokens & 77.67 / 4.18 & 77.35 / 4.31 & 77.06\\
\hline
\end{tabular}
}
}
\end{table}

\noindent\textbf{Vs. pooling:}
We compared against different pooling approaches similar to the ones tried in prior works, when they are required to select a much smaller number of tokens (e.g., 32) from a large set of visual tokens. We compare our spatio-temporal attentional pooling as well as the sequential model against its alternatives, including (1) fixed-window (non-learnable) space-time pooling and (2) learnable `per-frame' pooling. In particular, (2) is similar to the approach taken in LLaMA-VID~\cite{li2024llamavid}, which independently selected a fixed number of tokens (e.g., 2) per frame. Table \ref{tab:ablation32} shows the results.

\noindent\textbf{Number of visual tokens:}
Table \ref{tab:ablation-tokens} explicitly compares the impact of having smaller visual tokens. We are able to observe that 32 visual tokens or more gives a reasonable video QA accuracy.

\begin{table}
\begin{minipage}{0.55\linewidth}
    \caption{Ablations on sequential models (128).}
    \label{tab:ablation-ttm}
    \vspace{-0.5em}
    \centering
    \small
    \def\arraystretch{1.1}  
    \setlength\tabcolsep{0.5em}  
    \scalebox{0.88}{
    \begin{tabular}{c||c|c|c}
    \hline
    Temporal encoder & MSVD-QA  & ActNet-QA & NExT-QA\\
    \hline
    Original TTM & 76.42 / 4.15 &	54.45 / 3.48 &	75.42\\
    TTM + time-stamp &	76.43 / 4.16 &	56.15 / 3.53 &	75.96\\
    TTM + grouped &	76.99 / 4.17 &	55.92 / 3.54 &	76.46\\
    Ours  &	77.86 / 4.20 &	56.66 / 3.56 &	77.07\\
    \hline
    \end{tabular}
    }
\end{minipage}
\hspace{0.01\linewidth}
\begin{minipage}{0.43\linewidth}
    \caption{Ablations on the number of frames}
    \label{tab:ablation-frames}
    \centering
    \small
    \def\arraystretch{1.1}  
    \setlength\tabcolsep{0.5em}  
    \scalebox{0.90}{
    \begin{tabular}{c|c||c|c}
    \hline
    \# frames & \# tokens & NExT-QA & ActNet-QA\\
    \hline
    8 frames & 32 tokens & 76.4 & 55.7 / 3.5\\
    8 frames & 128 tokens & 77.1 & 56.7 / 3.6\\
    16 frames & 32 tokens & 76.7 & 55.9 / 3.5\\
    16 frames & 128 tokens & 77.6 & 57.3 / 3.6\\
    \hline
    \end{tabular}
    }
\end{minipage}
\end{table}

\noindent\textbf{Grouped token sequential model:}
Our grouped token sequential model temporal encoder implementation extends the conventional Token Turing Machines. 
Table \ref{tab:ablation-ttm} shows the ablations confirming the benefits of the new grouped token sequential model formulation. As we observe, the original TTM performs poorly. Adding the time-stamped positional encoding and memory grouping in the sequential model enables much better results, confirming their importance in the real-world models.


\noindent\textbf{More frames:}
In this experiment we validate whether \modelname is able to scale when trained to take more frames. Table \ref{tab:ablation-frames} shows the results. Even while maintaining the number of tokens, we are able to observe that providing more frames in the input allows \modelname to scale to better performance. We believe this is due to the fact that increasing the number of frames has an effect of increasing the size of the ``pool'' of tokens the temporal encoder can select from. We believe this trend will continue until it saturates.

\noindent\textbf{Speed:} Reducing the number of visual tokens increases the computational efficiency of the models, as the total computation is quadratic to the number of tokens fed to the LLM. We measure the runtime of our models in the training setting for the fair comparison.  Here, we report `samples per second per GPU'. Without the temporal encoder (i.e., directly using 1024 visual tokens), the model processed 3.3 samples per second. With 16/32/128 tokens using the temporal encoder, the model was able to process 8.5 / 8.2 / 7.5 samples per second.

\begin{figure*}
    \centering
    \includegraphics[width=0.85\linewidth]{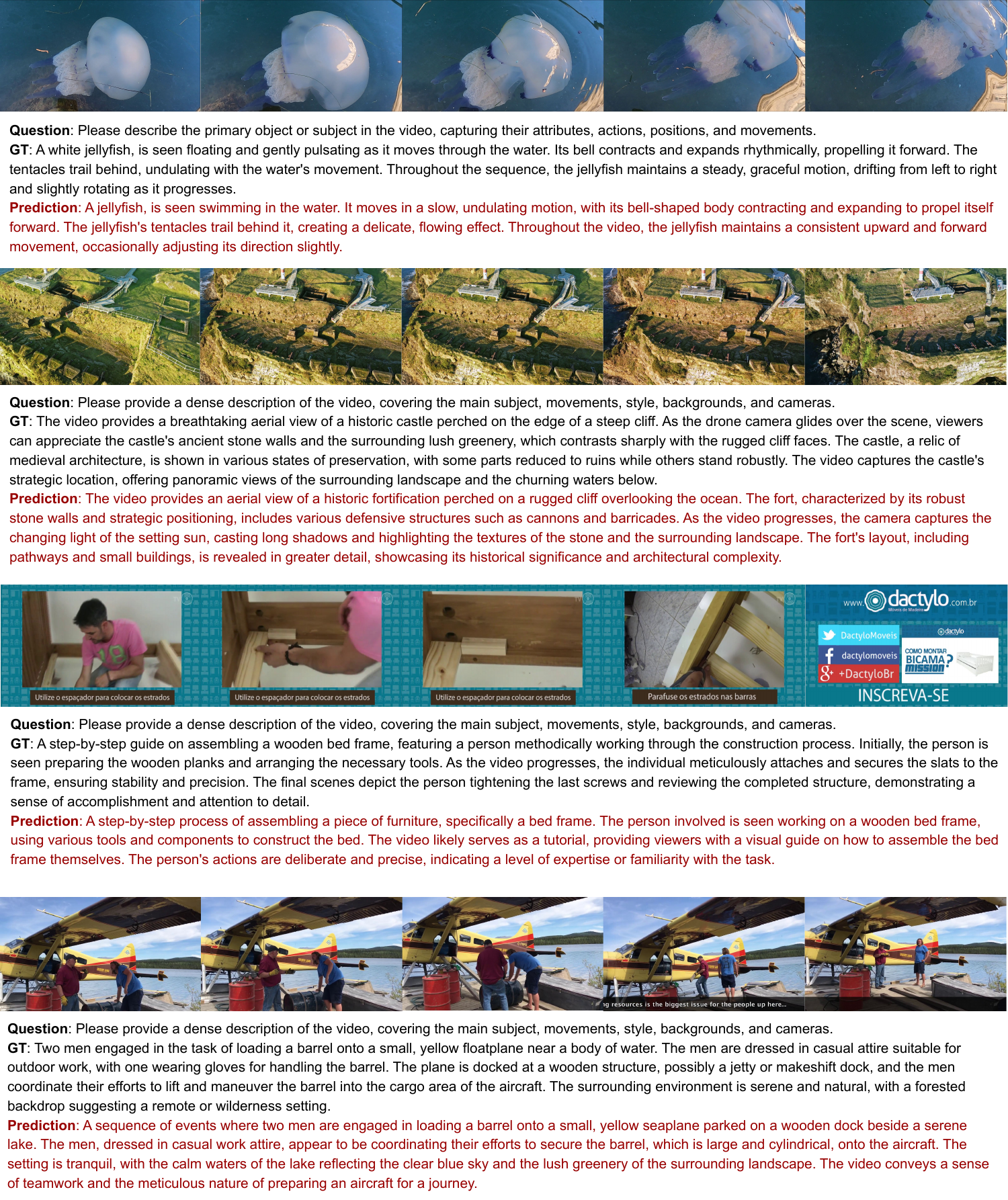}
    \vspace{-10pt}
    \caption{Example video captioning results on Mira dataset, formed in question-answering style.}
    \label{fig:caption_examples}
    \vspace{-5pt}
\end{figure*}

\begin{table*}[tb]
\centering
\caption{Video caption evaluation results using 8 frames. We employ VideoChatGPT's LLM evaluation and report Average Accuracy / Average Score. The `captioning-only model' was trained only using Mira video caption data (without QA data), making it specialized for the captioning task.}
{\small
\begin{tabular}{c|c|c||c|c|c}
\hline
Method & Size & \# tokens & MSVD-Cap & MSRVTT-Cap & Mira-Cap \\
\cline{2-4}
\hline
LLaVA-OneVision & 7B  & 1152 & 61.62 / 3.31  & 38.60 / 2.71   & 48.83 / 3.10 \\
Tarsier & 7B & 4608 &
 62.26 / 3.37  & 40.27 / 2.77 & 40.55 / 2.87 \\
\hline
\modelname & 4B & 32  & 63.59 / 3.38  & 42.06 / 2.82 & 80.67 / 3.96 \\
\modelname & 4B & 128  & 64.17 / 3.41  & 43.05 / 2.85  & 81.13 / 3.97 \\
\modelname (captioning-only model) & 4B & 128  & 69.50 / 3.52  & 50.45 / 2.98  & 81.76 / 4.00 \\
\hline
\end{tabular}
}
\label{tab:videocaption}
\end{table*}

\subsection{Video captioning evaluation}

We evaluate our model on the video captioning task by comparing it against state-of-the-art models on the test splits of MSVD-Caption and MSRVTT-Caption, as well as a custom evaluation split from the Mira dataset. For the Mira dataset, we randomly selected 6,000 samples from our full, filtered data to create the evaluation split, with the remainder used for training. We employed Video-ChatGPT's LLM evaluation, specifically using GPT-3.5 to compare model-predicted captions with ground truth captions. The LLM assesses accuracy by checking if the predicted caption matches the ground truth, and assigns a score on a scale of 0 to 5 for each sample.

Table~\ref{tab:videocaption} demonstrates the results. All three models were provided with 8 frames per video, and consistent visual input and prompts were ensured across the models. Our \modelname consistently outperforms LLaVA-OneVision-7B and Tarsier-7B across all three video captioning benchmarks, with particularly notable improvements on the Mira video captioning task.

We present qualitative video captioning results for the Mira dataset in Figure~\ref{fig:caption_examples} (and more in Appendix). \modelname generates high-quality, detailed captions.

\subsection{Longer video datasets}

We conducted experiments to confirm \modelname's capability to help long video understanding. We combined \modelname's captioning with a hierarchical text-based video understanding mechanism (i.e., LVNet \cite{Park2024TooMF}), and evaluated it on public benchmarks including EgoSchema and VideoMME. We use LVNet \cite{Park2024TooMF} as our backbone, relying on its key frame selection by extracting 8 frames centered at those key frames. \modelname was applied to such 8-frame segments for each key frame, serving as the video clip captioning model, and the generated captions were given to the high-level LLM in the LVNet framework.

Tables \ref{tab:videomme} and \ref{tab:egoschema} compares the results. As we observe, compared to the backbone framework used (LVNet), \modelname meaningfully improves its accuracy. It captures video information in the short intervals with LVNet, thereby enabling better question answering with long videos. Note that despite \modelname was not trained with long videos, it shows promising results. 

\begin{table}[tb]
\parbox{.49\linewidth}{
\caption{VideoMME long split results. 
}
\label{tab:videomme}
\centering
\small
\scalebox{1.00}{
\begin{tabular}{l|c}
\hline
Method          & Accuracy (\%) \\
\hline
LongVILA  & 38.8 \\
VideoChat-T   & 41.9   \\
LLaVA-OneVision & 43.8 \\
LLaVA-NexT-Video & 44.3 \\
VideoAgent      & 46.4   \\
Frame-Voyager  & 51.2   \\
VideoTree      & 53.1   \\
\hline
LVNet          & 52.4   \\
LVNet + BLIP-3-Video & 55.6   \\ 
\hline
LLaVA-Video  & 61.5 \\
InternVL2.5 & 62.6 \\
Qwen2-VL      & 62.2 \\
\hline
\end{tabular}
}
}
\hspace{0.02\linewidth}
\parbox{.49\linewidth}{
\caption{EgoSchema subset results. LVNet* accuracy is obtained by replicating the LVNet results by running its code with GPT-4o as LLM.}
\label{tab:egoschema}
\centering
\small
\scalebox{1.00}{
\begin{tabular}{l|c}
\hline
Method               & Accuracy (\%)  \\
\hline
LLoVi                & 57.6           \\
VideoAgent           & 60.2           \\
VideoTree            & 66.2           \\
\hline
LVNet*               & 66.2           \\
LVNet* + BLIP-3-Video & 67.4           \\
\hline
LifelongMemory       & 68.0           \\
Tarsier              & 68.6           \\
\hline
\end{tabular}
}
}
\end{table}

\section{Related Work}


\noindent\textbf{Image-Text LLMs.}
Recent image-text multimodal models~\cite{blip2,alayrac2022flamingo,liu2023llava,instructblip,xue2024xgenmmblip3,idefics3,molmo} typically use a pre-trained image encoder (e.g., ViT~\cite{openaiclip,siglip}) and a language-only LLM~\cite{abdin2024phi3,bai2023qwentechnicalreport,dubey2024llama3herdmodels}, connected via a vision-language connector that maps visual embeddings into LLM-compatible tokens. These tokens match the shape of language embeddings, allowing joint training via next-token prediction.
Connector designs vary: BLIP-2~\cite{blip2} uses a Q-Former, Flamingo~\cite{alayrac2022flamingo} uses a perceiver resampler and cross-attention, and others simply adopt MLPs \cite{liu2023llava}. 
Training is often multi-stage—pre-training, instruction tuning, and post-training (e.g., DPO~\cite{rafailov2024direct})—on both structured (e.g., captioning, VQA) and unstructured data (e.g., interleaved image-text~\cite{obelics,awadalla2024mint1tscalingopensourcemultimodal}, multi-image VQA~\cite{Jiang2024MANTISIM,li2024llavaonevision}).

\noindent\textbf{Video LLMs.}
Video LLMs extend image-based LLMs to handle video input. \cite{zhang2023video} uses frozen encoders and LLMs with Video/Audio Q-Formers for modality alignment. \cite{Maaz2023VideoChatGPT} extends CLIP with temporal features and fine-tunes on video-instruction pairs collected via tools like BLIP-2~\cite{blip2} and GRiT~\cite{wu2022grit}. \cite{li2024llamavid} compresses frame-level features into two tokens per frame but lacks temporal recency modeling. Video-LLaVA~\cite{lin2023video} and LLaVa-OneVision~\cite{li2024llavaonevision} treat videos as multi-image sequences, sacrificing compute efficiency. SlowFast-LLaVA~\cite{xu2024slowfast} introduces dual slow-fast pathways without extra fine-tuning. LLaVa-hound-DPO~\cite{zhang2024direct} explores DPO~\cite{rafailov2024direct} with GPT-4V-generated preference data. Our \modelname explores the orthogonal direction of efficiency via reduced tokens.

\noindent\textbf{Token Pruning.}
Token pruning reduces redundancy in ViTs and LLMs. \cite{bolya2022token, ren2023testa} merges similar tokens within ViTs.  \cite{shen2024tempme} focus on temporal redundancy and progressively merges tokens across neighboring clips.
\cite{chen2024image} prunes or merges tokens in deeper layers based on attention scores. \cite{shang2024llava} proposes adaptive token reduction via important token selection and supplementation. In LLMs, KV cache pruning improves efficiency~\cite{fu2024lazyllm}, and \cite{wan2024look} extends this to VLMs with token merging strategies. LongVU~\cite{shen2024longvu} spatially compresses frame tokens conditioned on the first frame, orthogonal to temporal reduction.
Our \modelname utilizes a novel temporal encoder to represent videos in as few as 16 or 32 tokens.

\section{Conclusion}
\label{sec:conclusion}
We introduce \modelname, which is an efficient, compact vision-language model for videos with 4B parameters. \modelname incorporates a temporal encoder in its architecture, which allows the model to abstract the entire video with as few as 16 or 32 tokens. In contrast to many state-of-the-art video VLMs taking advantage of thousands of visual tokens to represent a video (e.g., 4608), \modelname shows a competitive performance while utilizing much fewer visual tokens (e.g., 32).


\bibliography{refs}
\bibliographystyle{Styles/icml2025}


\newpage
\appendix
\section{Appendix}

\subsection{More qualitative examples}

\begin{figure}[ht]
    \centering
    \includegraphics[width=0.9\linewidth]{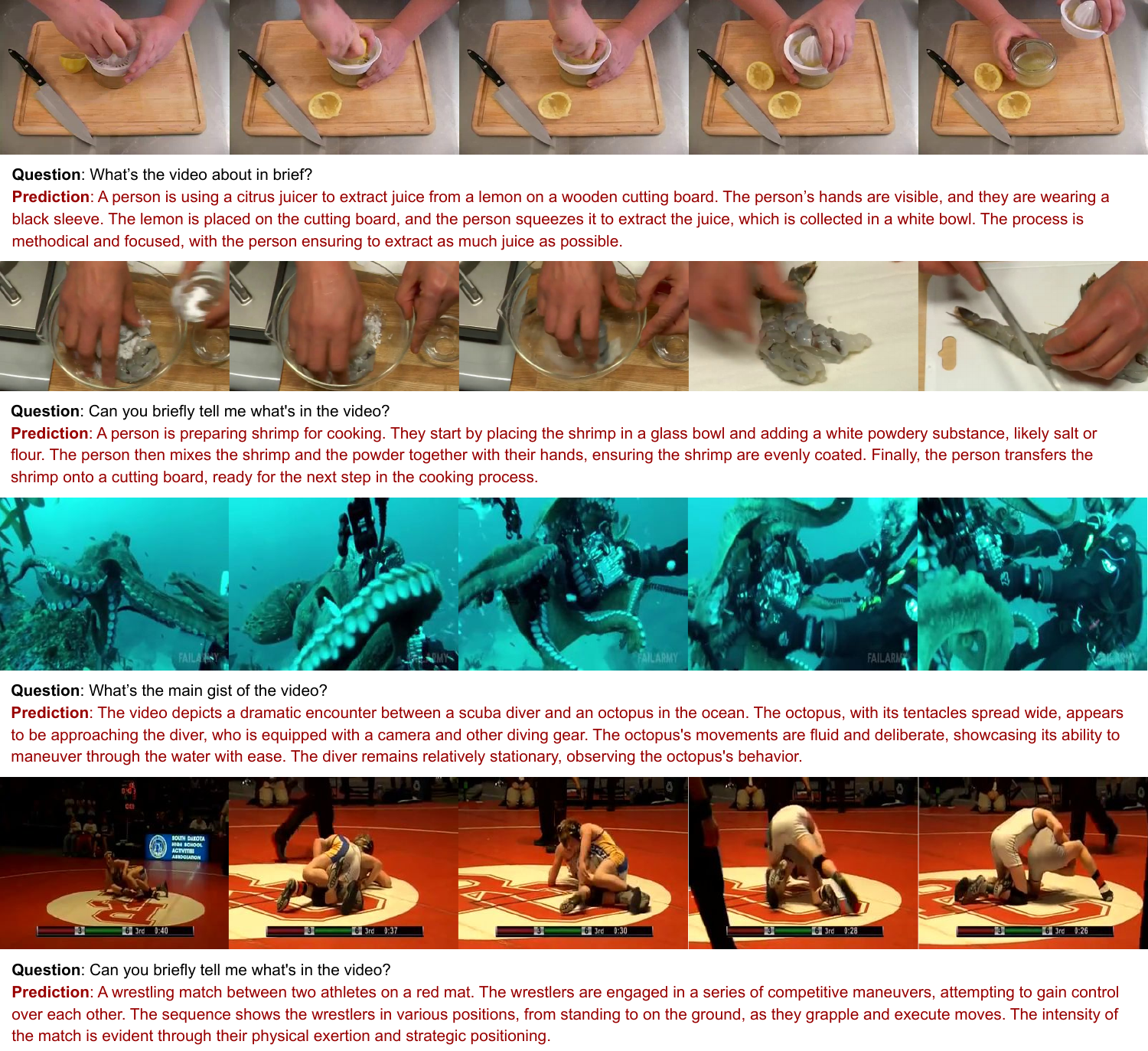}
    \caption{Example video captioning results on MSVD and MSRVTT caption dataset.}
    \label{fig:caption_examples2}
\end{figure}

Figure~\ref{fig:caption_examples2} shows \modelname's captioning results on the MSVD and MSRVTT datasets. We are able to observe that it generates reasonable captions despite utilizing a small number of visual tokens, e.g., 32.

\begin{figure*}
    \centering
    \includegraphics[width=0.99\linewidth]{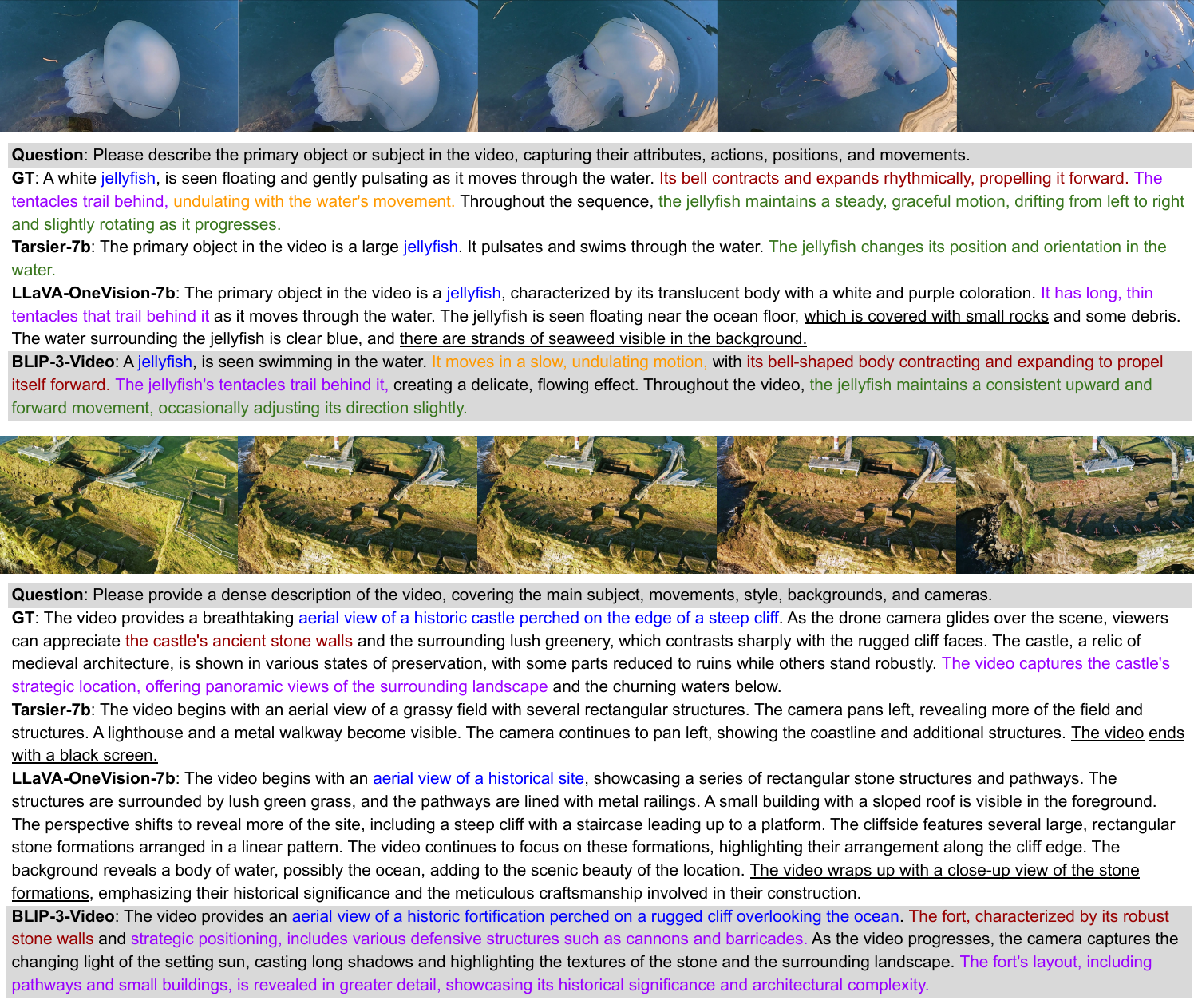}
    \caption{Example video captioning results on Mira dataset, formed in question-answering style. We compare the outputs of \modelname, Tarsier, and LLaVA-OneVision. GT stands for the ground truth. Different colored texts are different parts of ground truth captions and their corresponding sentences in the model outputs. Underlined texts are hallucinations.}
    \label{fig:caption_examples_details1}
\end{figure*}

\begin{figure*}
    \centering
    \includegraphics[width=0.99\linewidth]{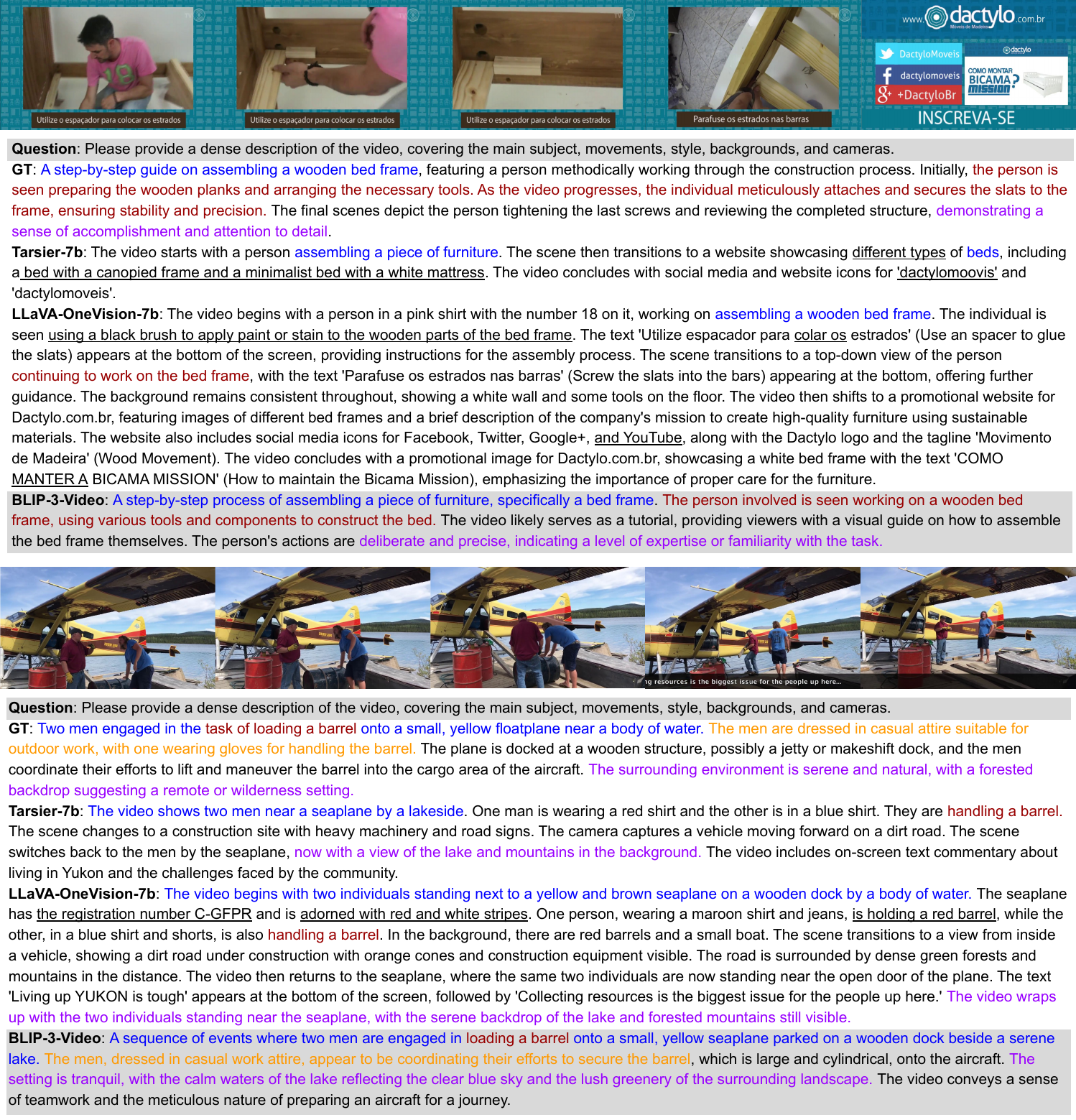}
    \caption{Example video captioning results on Mira dataset, formed in question-answering style. We compare the outputs of \modelname, Tarsier, and LLaVA-OneVision. GT stands for the ground truth. Different colored texts are different parts of ground truth captions and their corresponding sentences in the model outputs. Underlined texts are hallucinations.}
    \label{fig:caption_examples_details2}
\end{figure*}

Figures \ref{fig:caption_examples_details1} and \ref{fig:caption_examples_details2} compares the video captioning results of \modelname against Tarsier and LLaVA-OneVision, in detail. The models take the exact same video input with the same number of frames, for a fair comparison. We observe that \modelname generate captioning much more similar to the ground truth. It also is able to provide more holistic understanding of the video (rather than focusing on unnecessary details such as texts in the background) while suffering from less hallucinations.

\subsection{More experimental results}

We tested our model on additional benchmarks: VideoInstruct (which was used in VideoChatGPT) and TempCompass. Table \ref{tab:videoinstruct} shows the results on the VideoInstruct benchmark. We believe \modelname performs very reasonably on all these benchmarks, considering its smaller size and its use of much fewer visual tokens. Similarly, Table \ref{tab:tempcompass} shows \modelname's results on `Yes/No QA' and `Caption matching' tasks in TempCompass. They show similar trends to the other experiments. \modelname is a compact model giving us a reasonable accuracy.

\begin{table}
\parbox{.35\linewidth}{
\caption{Results on VideoInstruct benchmark.}
\label{tab:videoinstruct}
\centering
{\small
\begin{tabular}{c|c}
\hline
Model & Accuracy \\
\hline
PLLaVA-34B & 3.32\\
SlowFast-LLaVA-34B & 3.32\\
VideoGPT+ & 3.28\\
ST-LLM-7B & 3.15\\
\modelname & 3.11\\
VideoChat2\_HD\_mistral & 3.10\\
LITA-13B & 3.04\\
LLaMA-VID-13B & 2.99\\
VideoChat2 & 2.98\\
LLaMA-VID-7B & 2.89\\
Video-ChatGPT & 2.38\\
\hline
\end{tabular}
}
}
\hspace{0.03\linewidth}
\parbox{.61\linewidth}{
\caption{Results on TempCompass benchmark.}
\label{tab:tempcompass}
\centering
{\small
\begin{tabular}{c||c|c}
\hline
Model & Yes/No QA & Caption matching\\
\hline
GPT-4o & 73.66 & 80.84\\
Qwen2-VL-7B-Instruct & 72.77 & 77.31\\
Gemini-1.5-pro & 70.32 & 77.45\\
LLaVA-OneVision-Qwen-2-7B & 69.67 & 73.79\\
LLaVA-NeXT-Video-32B-Qwen & 69.38 & 76.51\\
InternVL2-8B & 68.24 & 77.11\\
\modelname & 66.7 & 66.5\\
Llama-3-VILA1.5-8B & 63.64 & 68.93\\
LongVA-7B & 62.13 & 65.67\\
LLaVA-NeXT-Video-7B-DPO & 61.19 & 63.01\\
VideoChat2-vicuna-stage3 & 58.01 & 53.69\\
LLaVA-1.5-13B & 56.38 & 64.27\\
Video-LLaVA-7B & 56.38 & 63.34\\
Video-LLaMA-2-13B & 53.73 & 54.16\\
LLaMA-VID-7B-short-video & 52.96 & 56.02\\
\hline
\end{tabular}
}}
\end{table}

\subsection{Reproducibility Statement}
\label{app:reproduce}

We build on top of the open-source BLIP-3 (XGen-MM) model and training code hosted on Huggingface and github. All the experiments were conducted with public datasets. The code and the trained model will be released together with the final version of the paper.

\subsection{Limitations}
\label{app:limitations}
At this point, all the models have been trained with videos up to a couple of minutes. Although we confirmed their capability to handle longer videos by combining it with a framework like LVNet, the model is expected to perform better with better training dataset curation with longer videos. The model sizes (3B) are also relatively small compared to other models. Despite our model showsing very promising model size-accuracy trade-off, training a bigger model with a larger dataset remains as future work.

\subsection{Societal Impact}
\label{app:impact}

Our proposed \modelname significantly improves video-language model efficiency by reducing the number of visual tokens needed question answering. This compact design lowers the computational and energy requirements, enabling more sustainable AI deployments and reducing the environmental footprint of large-scale video understanding. Furthermore, the compact architecture makes high-quality video-language reasoning more accessible to researchers and developers in resource-constrained settings, especially within academia. 

As with many powerful AI models, \modelname could be misused in ways that raise ethical concerns. The ability to process and understand video content efficiently may be exploited for mass surveillance, privacy-invasive monitoring, or unauthorized profiling. Moreover, if trained on biased or unbalanced data, the temporal encoder may overlook or misrepresent minority-relevant visual cues, potentially reinforcing harmful stereotypes. The model’s compact design could also facilitate easier generation harmful content or misinformation.

\end{document}